# Dilated CNNs for Periodic Signal Processing: A Low-Complexity Approach

Eli Gildish[a], Michael Grebshtein[a], Igor Makienko[a]

[a]*RSL Electronics LTD, Migdal Ha'Emek, 23100, Israel*

**Abstract**

Denoising of periodic signals and accurate waveform estimation are core tasks across many signal processing domains, including speech, music, medical diagnostics, radio, and sonar. Although deep learning methods have recently shown performance improvements over classical approaches, they require substantial computational resources and are usually trained separately for each signal observation. This study proposes a computationally efficient method based on DCNN and Re-sampling, termed R-DCNN, designed for operation under strict power and resource constraints. The approach targets signals with varying fundamental frequencies and requires only a single observation for training. It generalizes to additional signals via a lightweight resampling step that aligns time scales in signals with different frequencies to re-use the same network weights. Despite its low computational complexity, R-DCNN achieves performance comparable to state-of-the-art classical methods, such as autoregressive (AR)-based techniques, as well as conventional DCNNs trained individually for each observation. This combination of efficiency and performance makes the proposed method particularly well suited for deployment in resource-constrained environments without sacrificing denoising or estimation accuracy.

## 1. Introduction

Deep learning models designed to work with images, video and text are less fitted for processing and analysis of periodical signals corrupted by noise, like speech, EEG/ECG, radar and sonar signals, seismic signals and many others. This study proposes a new deep learning method fitted for periodical signals and optimized for minimum computational complexity and storage to be used in power limited devices like sensors or IoT devices. The method matches the performance of classical deep learning approaches with significantly reduced computational complexity. The proposed training-inference architecture requires a tiny network with very small number of parameters whose weights are trained on a single observation only and further used during inference.

One-dimensional (1D) Dilated Convolutional Neural Networks (DCNNs) have emerged as a cornerstone for processing time-series and sequential data. Unlike standard convolutions, dilated convolutions introduce gaps between kernel elements, enabling the network to capture a significantly larger receptive field without increasing the number of parameters or loosing resolution through pooling [1]. These networks are widely used in signal processing, denoising, prediction and other applications.

In signal processing they handle long-range temporal dependencies like: receptive field expansion to observe long segments of a signal where dilated filters allow a shallow network to incorporate more contextual information from the input signal [2]; multi-scale analysis of signals replacing the combination of discrete wavelet transforms with CNNs to allow simultaneous learning of long-term and short-term features in a single layer for biomedical signals like ECG and EEG [6]; improving efficiency compared to RNNs or LSTMs, since DCNNs are more computationally efficient (for example, in electrical power prediction it was 84% faster compared with LSTM [8]).

In denoising the use of multi-scale dilated convolution layers captures noise features across different temporal resolutions. This allows the network to adaptively enhance key signal channels while suppressing unstructured background noise [5]. DCNNs are frequently integrated with other techniques, such as Wavelet Transforms or autoencoders, to improve robustness. For example, in induction motor fault diagnosis, signals are preprocessed using autoencoders before entering a refined CNN architecture to handle complex, non-linear noise correlations. In high-noise environments, such as monitoring bearing faults, DCNN extracts features from both time and frequency domains simultaneously to ensure reliable classification despite significant disturbances [3] and provides high performance in vibration sources separation [9].

The predictive capabilities of DCNNs are widely utilized in biomedical applications like for seizure prediction and arrhythmia detection. For example, a combination of DCNNs with bidirectional RNN have achieved 99.9% accuracy in identifying cardiac irregularities [6]. In addition, DCNNs is a more stable alternative in time series forecasting compared with traditional LSTM showing less variability in results across different parameter configurations [4]. Dilated architectures are used in Speech Emotion Recognition (SER) to learn hidden local features and long-term contextual correlations from raw audio files, often outperforming deeper, more memory-intensive models [7].

Current DCNNs generally use fixed, integer-based dilation factors (e.g., $2^{l-1}$) for each layer. This constraint stems from the discrete nature of digitized signals, where signal period is constant. While this simplifies implementation during training, it significantly limits the network's flexibility to work with signals whose period changes like in most real-life applications. In real-world scenarios, when signal period varies the frequencies of signal components are not integer multiples of the sampling rate times dilation factor. Then the fixed dilation limits the network from achieving optimal performance due to the wrong temporal alignment. To compensate for it, existing DCNN use either non-linear activation functions with substantial increase in number of parameters or continuous re-training for every signal. The solutions significantly degrade convergence and increase computational overhead limiting the use of DCNN in power-limited edge applications.

This paper proposes a novel method of using DCNN with Re-sampling (R-DCNN) for de-noising of periodical signals whose period varies. The study proves that the denoising can be performed without re-training on new observations when their fundamental frequency is known. The model is trained by using a single observation only tested on re-sampled observations whose time axis is transformed to match the time axis of the trading observation. The paper proves that when signal fundamental frequency varies the only parameter which should be adjusted is time scaling allowing to use the model with fixed weights.

The experiments demonstrate that the new R-DCNN provides high accuracy of signal denoising while significantly decreasing computational complexity compared to the existing methods including deep learning DCNN and even classical autoregression (AR) in the estimation of periodic signals with varying period.

2. **Problem Statement**

Consider a discrete-time stochastic process $y[n]$ observed over a window $N$:

$$y[n] = x[n] + \eta[n], \quad n \in [0, N-1] \qquad (1)$$

Where:

- $x[n]$ is a deterministic periodic process: $x[n] = x[n + P_0]$ for some period $P_0$ (in samples). $P = f_s/F_0$, where $f_s$ is the sampling frequency. The signal is composed of sinusoids some of them correspond to fundamental frequency $F_0$ and its harmonics.

- In real life $\eta[n]$ is an additive color noise such that $\eta[n] = \sum_{j=1}^{J} b_j \varepsilon[n-j]$ and $\varepsilon[n]$ is uncorrelated innovation process.

The goal is to estimate deterministic periodic process $\hat{x}[n]$ given that its fundamental frequency $F_0$ is known. Minimum computational complexity is required to allow the method to be used in edge devices.

## 3. Methodology

*3.1. Existence of the Optimal and Unique Solution*

Since the noise is colored the estimation should be done by separating between the internal signal dependencies inside the periodical and the stochastic noise with correlation. According to Wold's theorem [12] it is always possible to achieve a unique decomposition of a process into deterministic (or periodic) and stochastic parts. In a periodic process $x[n]$, any observation can be reconstructed without error by using a linear combination of its past observations $x[n-i]$, where $i > 0$. However, in stochastic process $\eta[n]$, prediction of its future values based on past values results in a random error.

For periodic signals the optimal predictor in terms of mean squared error is a linear combination of its past values. The predictor for values in $y[n]$ is equivalent to the estimator of the periodic signal $\hat{x}[n]$ as follows:

$$\hat{y}[n + \Delta] = \hat{x}[n + \Delta] = \sum_{m=1}^{M} w_m \cdot y[n - m] \qquad (2)$$

where the predictor $\hat{x}[n]$ estimates the value $y[n]$ by utilizing its previous $M$ values given a specific value of $F_0$. According to [10] the time delay $\Delta$ guarantees that the stochastic component remains uncorrelated, as expressed by the condition $E\{\eta[n]\eta[n-m]\} = 0, \forall\, m > \Delta$.

For the sake of clarity, the delay will be omitted in the following sections and will be taken into consideration only in the implementation for predicting $\hat{y}[n + \Delta]$ instead of $\hat{y}[n]$.

Once estimated, the solution for $w_m$ is unique and optimal for a deterministic process, as proven by Wold. Thus, the estimator of $\hat{x}$ exist and unique given the coefficients $w_m$ and delay $\Delta$.

*3.2. Problem Re-Formulation*

In practice, the fundamental frequency of a periodic signal varies over time. Consider a reference signal with a known fundamental frequency $F_{ref}$. The fundamental frequency of another signal can then be expressed as $F_i = s_i F_{ref}$, where $s_i$ is a scaling factor (equivalently, the period scales by $1/s_i$).

For the resulting signal, denoising can be performed using the same coefficients $w_m$, applied to the values $\tilde{y}_i[n/s_i]$. Since these values generally do not lie on the discrete-time grid, they must be obtained via interpolation.

The same set of weights $w_m$ can also be used to estimate any periodic deterministic process $\hat{x}_i[n]$ after scaling the reference frequency by $s_i$. In this case, the time axis of the corresponding observation must be scaled by $1/s_i$ to match that of the reference signal.

$$\hat{x}_{ref}[n] = \sum_{m=1}^{M} w_m \cdot y_{ref}[n - m]$$
$$\hat{x}_i[n/s_i] = \sum_{m=1}^{M} w_m \cdot \tilde{y}_i[n/s_i - m/s_i] \qquad (3)$$

Where:
- $\tilde{y}$ represents the interpolated observation $y[n]$ at fractional indices depending on $s_i$

This means that, once estimated, the weights may be fixed when the fundamental frequency changes and the scaling factor (or fundamental frequency) is known.

### 3.3. Periodic Signal Denoising

There are many methods for periodic signal denoising, which can be broadly categorized into parametric, nonparametric, and more recently, machine learning (ML) - based approaches. Among the nonparametric (spectral) methods, the Lomb–Scargle periodogram (LSP) is widely used to estimate spectral density via a least-squares fit of sinusoids. For practical applications, study [14] extends LSP to also estimate phase and amplitude spectra. Study [15] introduces SigSpec, which derives an analytical probability density for amplitude in frequency domain under white noise, incorporating both frequency and phase. It is reported to be less susceptible to aliasing than LSP or Phase Dispersion Minimization. A comparative study [16] benchmarks several nonparametric frequency estimators in terms of bias, standard deviation, and computational complexity.

Parametric and model-based methods approach the problem differently. Study [17] surveys parametric frequency estimation for sinusoids in noise, covering autoregressive (AR) modeling via Yule–Walker equations, high-resolution subspace methods (Root-MUSIC, ESPRIT, truncated SVD), maximum likelihood, and Bayesian approaches, along with the critical issue of model order selection.

Adaptive and filtering methods address signals with slowly varying frequencies. Paper [18] proposes an adaptive algorithm based on discounted least squares for enhancement and tracking of periodic signals with fixed or slowly varying frequency. Study [19] derives a nonlinear observer for estimating parameters of multi-sinusoidal (possibly harmonically related) signals, reporting improved performance over the extended Kalman filter. In [20] the authors present a nonlinear adaptive time-domain method for instantaneous frequency estimation with strong noise immunity.

Sparse and compressed sensing methods represent a more recent direction. Study [21] proposes recovery of periodic mixtures via sparse representations in nested periodic dictionaries, deriving improved support recovery guarantees using newly introduced coherence measures. In [22], the authors show that exploiting the periodic support structure of clustered sparse signals improves upon classical compressed sensing bounds.

Machine learning - based methods can be divided into two groups: learning-based frequency estimation and periodicity - exploiting denoising.

On the frequency estimation side, a notable contribution is [23], presented at NeurIPS, which proposes a neural network architecture that outputs a learned representation with local maxima corresponding to estimated frequencies of multi-sinusoidal signals. The model includes a dedicated module for estimating the number of frequency components and achieves state-of-the-art performance, outperforming existing techniques "by a substantial margin at medium-to-high noise levels." In [24], the authors propose a hybrid approach: a DnCNN (residual learning denoising CNN) is first used to remove unstructured noise, followed by a conventional subspace estimator (e.g., MUSIC/ESPRIT) applied to the denoised signal. This two-stage approach outperforms atomic norm minimization–based denoising and significantly improves line spectral estimation compared to applying subspace methods directly to noisy data. Study [25] demonstrates a simpler three-layer neural network for single-frequency estimation from noisy sinusoids at 25 dB SNR, achieving very low mean squared error for normalized frequencies with inference times under one second.

For periodicity-exploiting denoising, [26] propose a framework specifically designed for periodic signals by folding a 1D waveform into a 2D grid (one period per row) and applying fully convolutional 2D denoisers. This allows the network to exploit cross-period correlations, conceptually similar to classical synchronous averaging but learned. A key advantage is that a single trained model generalizes across a range of frequencies, including unseen ones. The same group in [27] applied WaveNet-based architectures to periodic signals with strong discontinuities, using a total variation cost function and training solely on synthetic data, with successful transfer to real friction signals from tribological systems. Similarly, [28] applies a 2D stacking approach to ECG denoising, where cardiac cycles are arranged into a 2D tensor and processed using a CNN with a local/non-local cycle observation module, demonstrating improvements on the MIT-BIH Arrhythmia database.

Learnable wavelet-based approaches were introduced starting with [29], which presents a learnable wavelet packet transform - a signal-processing-inspired deep architecture with few parameters and intuitive initialization. It demonstrates strong denoising performance across signal classes unseen during training and robustness to varying noise levels. In [30], a Dual-Domain Denoising Network (D3N) is proposed, integrating trainable lifting schemes with a dual-path architecture: a recurrent attention branch for global temporal dependencies and a sparse wavelet coding branch for local spectral features. Significant improvements over baseline methods are reported on ECG and seismic datasets. In [31], a fast wavelet transform is combined with a feedforward network that learns binary weights for sub-band selection, effectively discarding noisy wavelet components during photoplethysmogram (PPG) signal reconstruction.

For representation learning of periodic time series, [32] proposes Floss, an unsupervised frequency-domain regularization method that detects dominant periodicities and enforces periodic consistency in learned representations via spectral density similarity measures. It can be integrated into existing deep learning models for classification, forecasting, and anomaly detection. Study [33] introduces a CNN-based period classification algorithm that learns period lengths from waveform shape rather than amplitude, achieving perfect accuracy in low-noise conditions and degrading gracefully with increasing noise. In [34], the authors argue that conventional preprocessing filters (e.g., Butterworth, FIR, wavelet) distort cyclic waveforms and degrade downstream deep learning performance; they propose a self-adaptive frequency decomposition method with regularization, reporting a 45% improvement in MSE over Butterworth filtering for PPG signals.

Several trends emerge from this literature. The most effective approaches for periodic signals explicitly exploit periodic structure—either by transforming 1D signals into 2D representations ([26]-[28]) or by combining learned denoising with classical spectral estimators ([24]). Pure end-to-end approaches such as [23] perform well for frequency estimation but do not necessarily reconstruct the full waveform. Learnable wavelet-based methods provide a promising balance between interpretability and flexibility.

Recent and rapidly developing approaches, including transformer-based models, physics-informed neural networks, and diffusion-based denoisers for periodic signals, are not covered in this survey.

Machine learning approaches typically require significant computational resources for both training and inference, which limits their applicability in power-constrained edge devices. Furthermore, many classical periodic signal denoising methods are not directly suitable for IoT applications, as they require parameter optimization for each observation.

As baseline methods for new developments, autoregressive (AR) modeling and self-adaptive noise cancellation (SANC) can be considered. These methods predict periodic signal components using short-term historical data, as proposed in [35] and further improved in [36] through multi-delay prediction to reduce variance. However, they assume long and stationary recordings to achieve high frequency resolution, making them less robust when the fundamental frequency varies.

More recently, studies [9] and [13] address this limitation using dilated convolutional neural networks (DCNNs), which improve frequency resolution by increasing the receptive field without increasing the number of trainable parameters. Nevertheless, a key drawback is that model parameters must be re-estimated for each new observation, limiting their practicality under strict power constraints.

In this work, a new method for periodic signal denoising using DCNNs under power constraints is developed and compared with existing state-of-the-art approaches.

*3.4. R-DCNN - Dilated CNN with Resampling*

The estimation procedure consists of two stages: training and inference. During the training stage, a 1D linear dilated convolutional neural network (DCNN) is employed instead of autoregressive (AR) - based methods, which typically require very large window sizes to achieve high frequency resolution. The DCNN is trained using a single reference observation, after which its architecture and parameters are fixed. During the inference stage, the trained model is applied to all observations. Prior to processing, each observation is resampled to ensure synchronization with the reference observation used during training.

The algorithm employs a linear one-dimensional DCNN to obtain an optimal linear predictor for the periodic signal, as shown in (3). The estimation is carried out such that each output sample $y[n + \Delta]$ is predicted using the preceding $M$ samples $y[n-1], \ldots, y[n-M]$.

The DCNN network parameters are:

$$\Theta = \{W^{(1)}, \dots, W^{(L)}\}$$

where $W^{(\ell)}$ are the convolution weights of layer $\ell$,

The dilation of layer $\ell$ is defined as:

$$d_\ell = 2^{\ell-1} \tag{4}$$

Let $h_r^{(\ell)}[n]$ denote the $r$-th channel of layer $\ell$. The input is

$$h^{(0)}[n] = y[n-1], \dots y[n-M]$$

For a kernel of length $K_\ell$, the dilated convolution of layer $\ell$ is calculated as follows:

$$h_r^{(\ell)}[n] = \sum_{c=1}^{C_{\ell-1}} \sum_{k=0}^{K_\ell - 1} w_{rck}^{(\ell)} h_c^{(\ell-1)}[n - k d_\ell] \tag{5}$$

Where $w_{rck}^{(\ell)}$ are weights of the $\ell$ layer corresponding to channel $r$.

An example of a 3-layer DCNN is shown in Figure 1

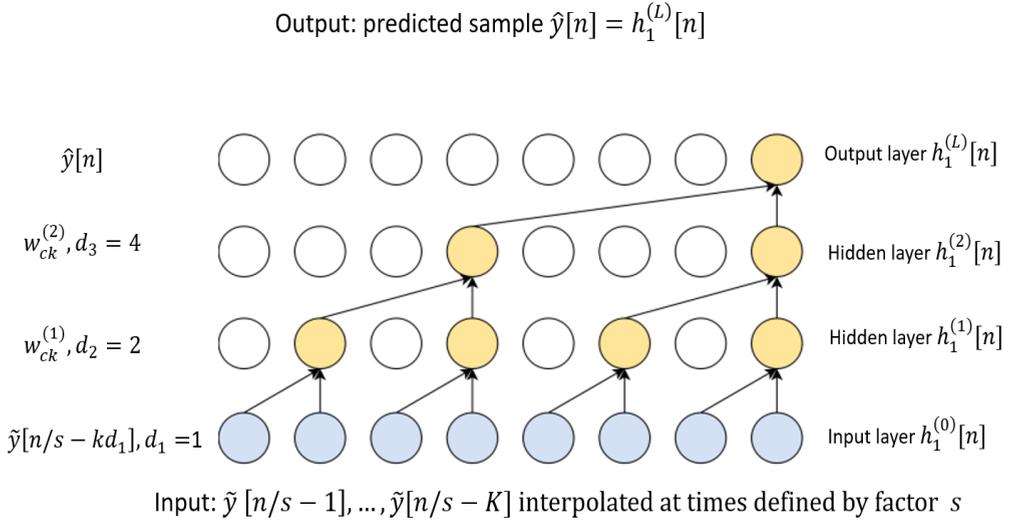

Figure 1. An example of a single-channel DCNN with two hidden layers and a scaling factor $s$

When signal fundamental frequency $F_0$ is scaled by factor $s$ the predictor changes as follows:

$$\hat{x}[n/s] = \sum_{k=1}^{K} w_k \cdot \tilde{y}[n/s - k/s] \tag{6}$$

This implies that the model weights $w_k$ may remain unchanged when the time-scaling $s$ varies with the fundamental frequency $F_0$, resulting in a resampled signal $\tilde{y}[n/s]$, obtained via interpolation at time instances $n/s$.

Accordingly, the DCNN model, whose parameters are learned during training from a single observation, can be applied to denoise any new observation, with a fundamental frequency $F$, after a resampling with a scaling factor $s = F/F_0$. The scaling factor $s$ is assumed to be known in this study; however, in practice, it can be estimated using low-complexity spectral methods, such as spectrum-based frequency estimation techniques like in [13].

The resampling can be implemented using a rational factor representation with upsampling factor $U$ and downsampling factor $L$, such that $s = U/L$. To ensure computational efficiency, the operation is performed using polyphase filtering in the downsampled domain, satisfying $s = U/L$ while minimizing computational complexity:

$$\text{int}(U, L > 0), \text{argmin}(U + L), \quad s.t. |s - U/L| \leq \varepsilon \tag{7}$$

The parameters $U$ and $L$ are determined using a standard rational approximation of the scaling factor $s$, based on the required precision, or by identifying a common denominator:

$$s = F/F_0, F > F_0$$
$$U = \frac{F}{\gcd(F, F_0)}, \quad L = \frac{F_0}{\gcd(F, F_0)} \tag{8}$$

The precision of $s$ determines the accuracy of the resampling process. Its selection is guided by the desired trade-off between denoising performance and computational complexity.

An alternative approach for determining $U$ and $L$ is based on a fractional approximation algorithm. In this case, the scaling factor $s$ is represented as a sequence of rational approximations $U_n/L_n$, which progressively converge to the true value of $s$, as follows:

$$s = a_0 + \cfrac{1}{a_1 + \cfrac{1}{a_2 + \cfrac{1}{a_3 + \cdots}}} \tag{9}$$

The approximation is obtained iteratively by extracting the integer part of $s$ and its remainder at each step. The remainder is then inverted, and the procedure is applied recursively. The corresponding numerator and denominator terms, $U$ and $L$, are updated at each iteration as follows:

$$a_n = \lfloor 1/r_{n-1} \rfloor$$
$$U_n = a_n U_{n-1} + U_{n-2}, \quad U_{-2} = 0, U_{-1} = 1 \tag{10}$$
$$L_n = a_n L_{n-1} + L_{n-2}, \quad L_{-2} = 1, L_{-1} = 0$$

where $a_n$ are the coefficients defined as the integer part of the inverted remainder from the previous iteration. The initialization is given by $a_0 = \lfloor s \rfloor$ and $r_0 = s - a_0$. At each iteration, the approximation accuracy improves; however, the corresponding values of $U$ and $L$ also increase, leading to higher computational complexity.

The resampling process using polyphase filtering is illustrated schematically in Figure 2:

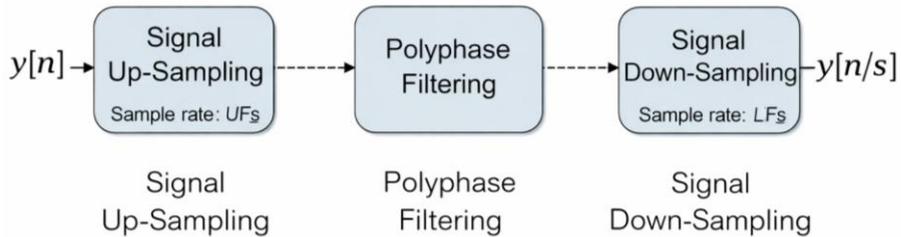

Figure 2. Signal re-sampling according to scale factor $U/L$

## 4. Algorithm Summary:

*4.1. R-DCNN Training for Weights Estimation*

A one-dimensional linear DCNN is used, with its weights optimized by minimizing the prediction error for a single reference observation:

$$\mathcal{L} = \frac{1}{2N} \sum_n (\hat{y}[n] - y[n])^2$$

*Required:*

A single training observation $y[n]$; number of DCNN layers $L$; kernel lengths $K_\ell$ for each layer; channel counts $C_\ell$; learning rate $\eta$, number of epochs, batch size.

*Initialization:*
Weights $w_{rck}^{(\ell)}$ of each layer with small random values; dilations $d = 2^{\ell-1}$

*Training:*
The network is trained on a single observation $y[n]$, during which the weights $w_{rck}^{(l)}$ are optimized. It is assumed that the signal fundamental frequency remains constant over the duration of this observation.

*4.2. Inference (R-DCNN)*

For each new observation with known fundamental frequency $F$, the scaling factor $s = F/F_0$ is first estimated. The upsampling and downsampling factors, $U$ and $L$, are then computed according to (9) - (10), and resampling is performed using a polyphase filter. The deterministic periodic signal $x[n]$ is subsequently estimated by applying the trained DCNN to the resampled observation. The stages of the algorithm are illustrated schematically in Figure 2.

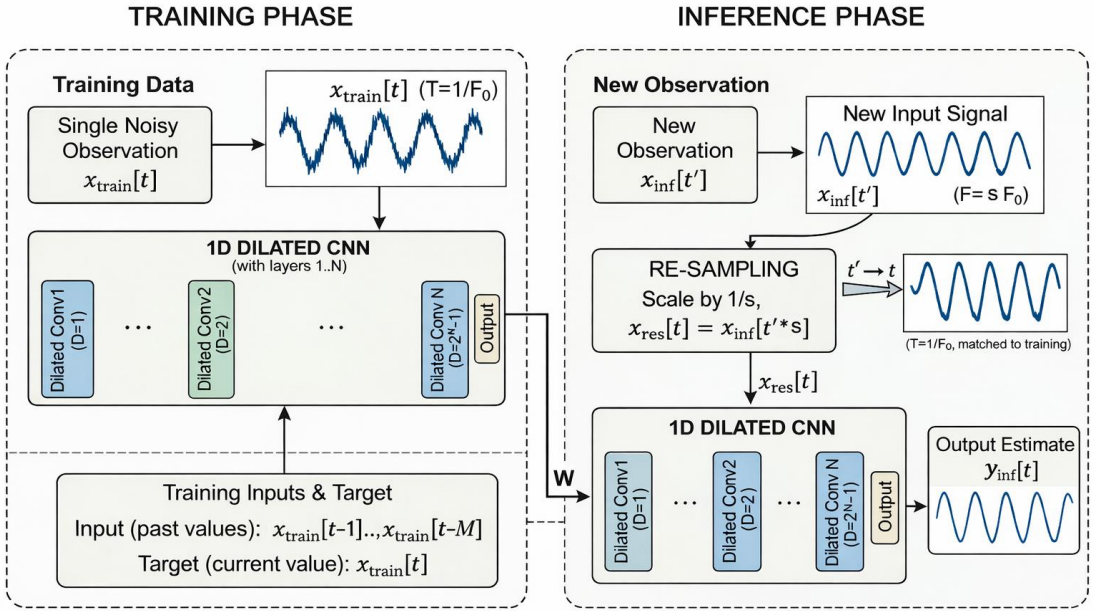

Figure 3. The proposed R-DCNN method for periodic signal denoising operates in two stages. First, a DCNN

is trained using a single reference observation with known fundamental frequency $F_0$. Second, the trained model is applied to new observations with fundamental frequency $F = sF_0$, followed by resampling to match their time scale to that of the reference signal.

## 5. Experimental Results

*5.1. Simulations*

The simulations were conducted to evaluate the accuracy of the proposed R-DCNN in comparison with existing approaches for periodic deterministic signal estimation based on historical samples, such as deep learning DCNN and classical autoregressive (AR) methods. These methods require retraining for each observation, leading to significant computational overhead. The objective is to demonstrate that the estimation accuracy of the R-DCNN is comparable to that of classical methods, thereby making it a more suitable solution for periodic signal denoising in power-constrained applications.

Three methods: R-DCNN, DCNN, and AR are evaluated by comparing the input SNR with the output SNR after denoising over the frequency $F_0$ range of 20 to 50 Hz. For each method, the best performance across the entire SNR range is considered. Two key parameters are varied in the simulations: the number of harmonics (5, 10, and 20) and the input SNR levels (-10, -5, 0, 10, and 15 dB).

The reported results are based on 100 independent realizations for each method and parameter configuration. In each realization, the deterministic signal $x[n]$ is generated as a sum of sinusoids comprising a fundamental frequency and its harmonics, with amplitudes and phases uniformly distributed over the ranges [0, 1] and $(0, \pi/2)$ respectively. The fundamental frequency varies between realizations. Each observation $y[n]$ is formed by adding white noise to $x[n]$, according to the specified SNR.

*5.2. Settings*

Sampling frequency: 2000 Hz

Signal length: 2 sec

Number of signals per iteration: 100

Number of iterations: 100

**AR:**
- The window length was varied over the range [20, 40, 60, 80, 100] samples to obtain different receptive fields and corresponding frequency resolutions.

**DCNN:**
- The model is retrained for each new observation.
- Architecture: [2, 3, 4, 5] layers with a kernel size of 8 samples. The total number of parameters ([16, 24, 32, 40]) is determined by the number of layers and kernel size, yielding receptive fields comparable to those of AR methods.

**R-DCNN:**
- Number of signals for training: 1; for inference: 99
- Architecture: the same as in DCNN
- Fundamental frequency $F_0$ in training observation and $F$ in inference observations are known and used to compute the scaling factor $s$
- Re-sampling precision during inference: $|s - U/L| \leq \varepsilon, \varepsilon = 0.1$

## 5.3. Results

The simulation results, presented in Figure 4, demonstrate that the performance of the proposed R-DCNN is comparable to that of existing methods. This indicates that R-DCNN is a favorable choice for periodic signal denoising in power-constrained applications. The evaluation is conducted across a range of SNRs, number of harmonics, and fundamental frequencies. The DCNN trained separately for each observation achieves the best performance; however, it incurs significantly higher computational complexity. In contrast, the average performance of R-DCNN and AR methods is similar, with R-DCNN showing an advantage at lower SNR levels. This advantage arises from the assumption that the fundamental frequency of each observation is known, enabling accurate resampling. In practical scenarios, the performance of R-DCNN may degrade due to errors in fundamental frequency estimation. However, this aspect is beyond the scope of the present study.

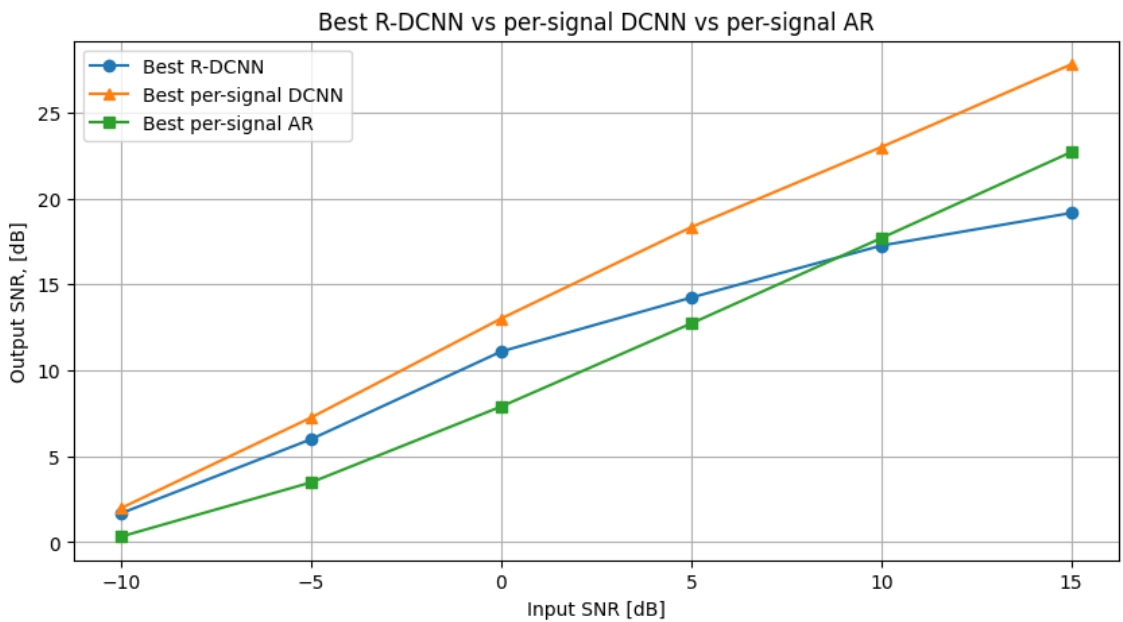

Figure 4. Performance comparison of the proposed R-DCNN with conventional DCNN and AR methods for periodic signal denoising. In contrast to the latter, R-DCNN eliminates the need for per-observation training. The x-axis denotes the input SNR, and the y-axis denotes the output SNR after denoising.

The performance of the R-DCNN for different numbers of layers is shown in Figure 6. The results indicate a high level of stability across configurations; however, the accuracy is lower compared to the per-observation DCNN presented in Figure 6. In both methods, performance degrades as the number of layers decreases, since the resulting receptive field becomes insufficient to achieve the required frequency resolution, particularly for signals composed of closely spaced sinusoids. Similarly, the AR performance, illustrated in Figure 7 exhibits degradation as the window length decreases. This is due to the same limitation in effective receptive field length observed in DCNN-based methods.

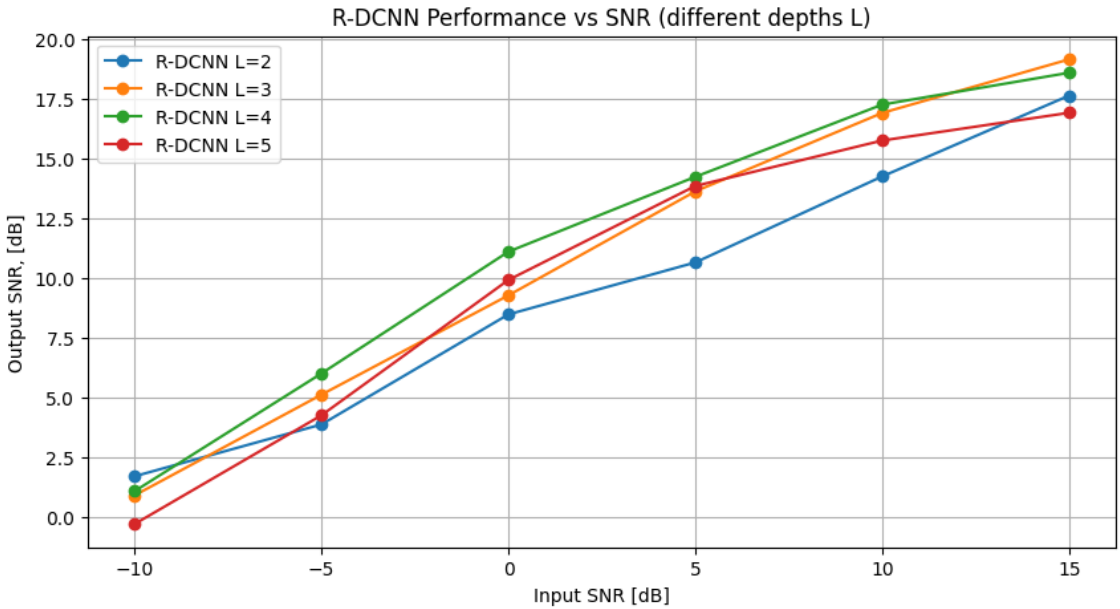

Figure 5. Performance of the R-DCNN trained on a single reference observation for different numbers of layers. The x-axis represents the input SNR of the noisy observation, and the y-axis represents the output SNR after denoising. The performance is consistent across different layer configurations, with underfitting observed for $L = 2$.

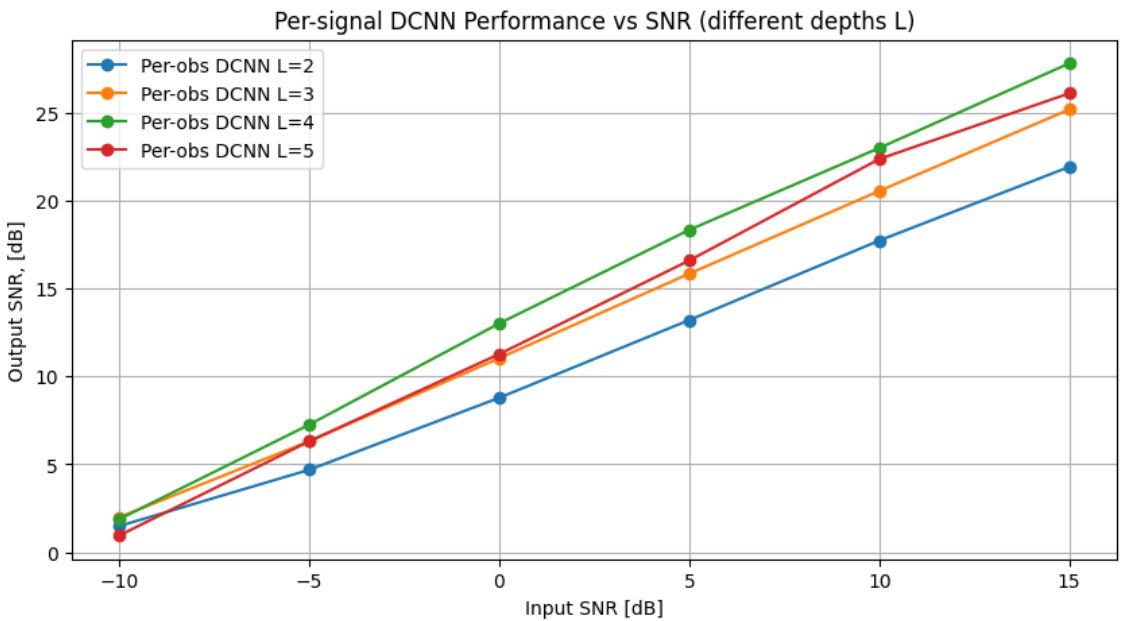

Figure 6. Performance of the per-observation DCNN for varying numbers of layers. The x-axis denotes the input SNR, and the y-axis denotes the output SNR after denoising. A noticeable performance degradation occurs at $L = 2$, due to the insufficient receptive field for capturing the required frequency resolution.

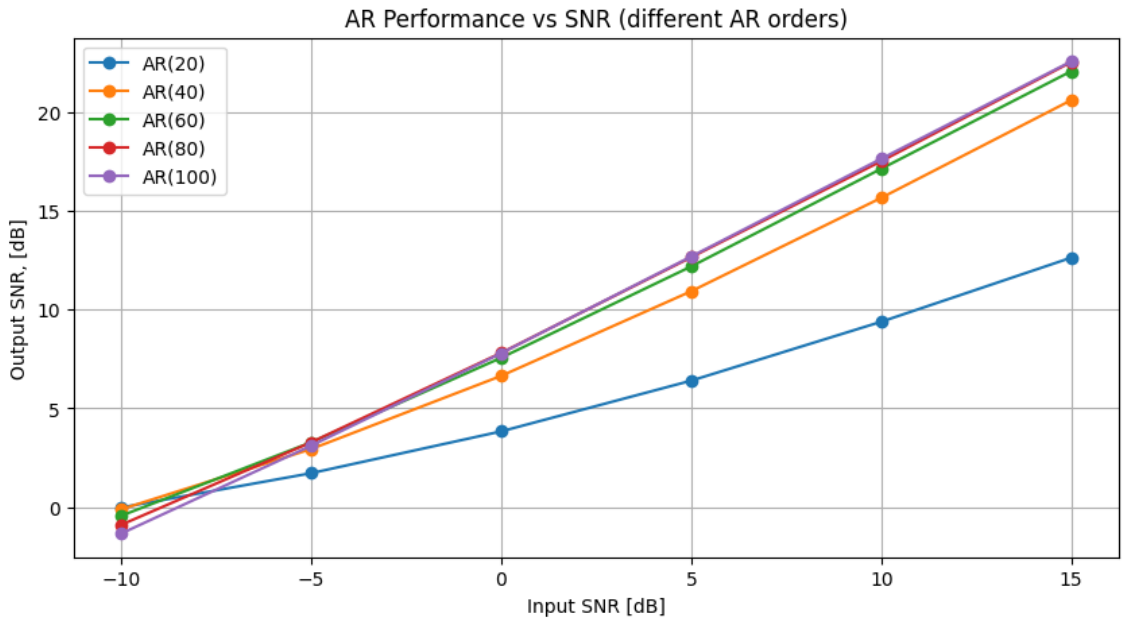

Figure 7. Performance of AR-based denoising with per-observation optimization for varying prediction window lengths. The x-axis denotes the input SNR, and the y-axis denotes the output SNR after denoising. Reduced performance is observed for shorter window lengths (20 and 40 samples), due to the insufficient receptive field for accurate frequency resolution.

Some examples of the periodic signal waveform estimation are shown in few figures below under different SNR conditions.

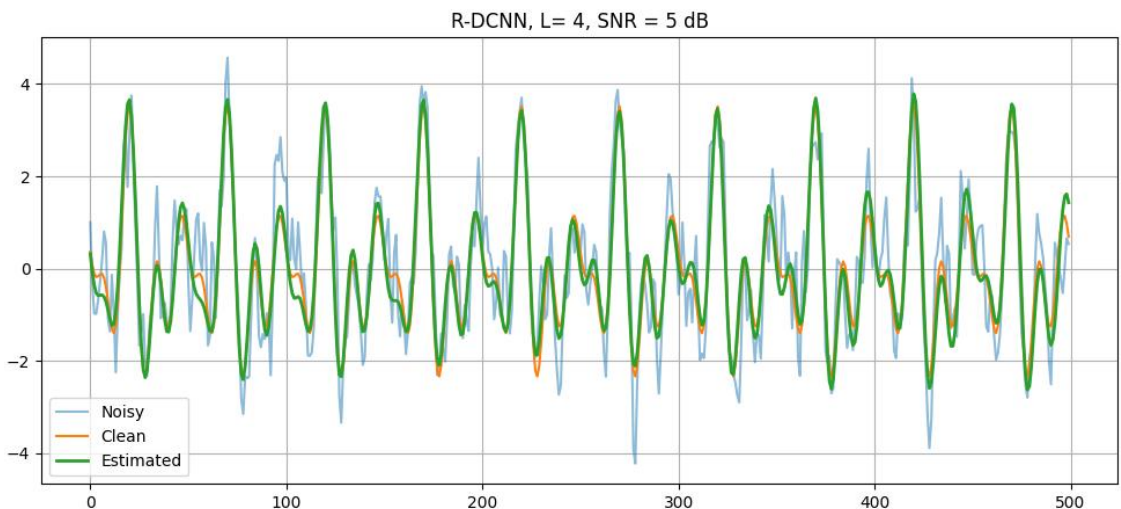

Figure 8. R-DCNN-based periodic signal waveform estimation at 5 dB SNR with $L = 4$ layers. The estimated signal (green) is shown alongside the clean signal (orange) and the noisy input signal (blue).

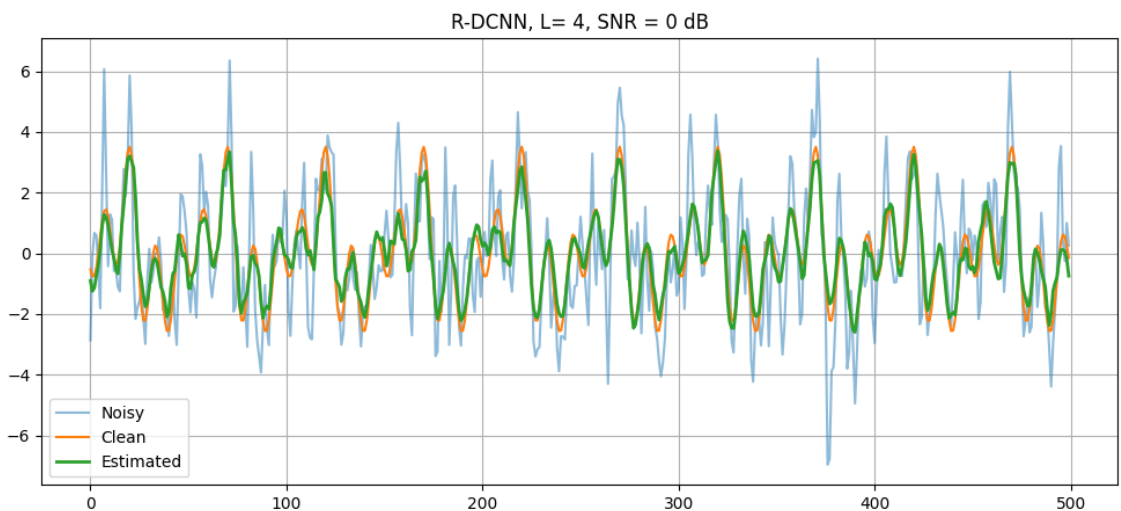

Figure 9. R-DCNN-based periodic signal waveform estimation at 0 dB SNR with $L = 4$ layers. The estimated signal (green) is shown alongside the clean signal (orange) and the noisy input signal (blue).

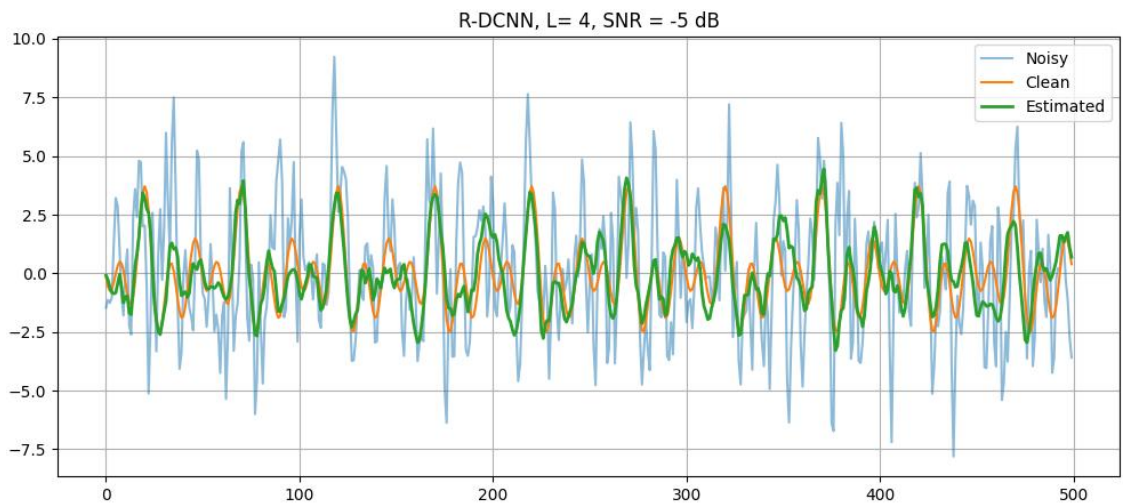

Figure 10. R-DCNN-based periodic signal waveform estimation at -5 dB SNR with $L = 4$ layers. The estimated signal (green) is shown alongside the clean signal (orange) and the noisy input signal (blue).

*5.4. Conclusions and Future Work*

This study presents a novel R-DCNN method for estimating periodic deterministic signals from noisy observations with varying fundamental frequency. On the one hand, the method leverages the advantages of DCNNs, enabling a large receptive field without increasing the number of parameters. On the other hand, it is adapted for low computational complexity, making it suitable for power-constrained applications such as edge IoT devices.

Unlike conventional approaches, which require retraining the DCNN for each observation, the proposed method trains the network on a single reference observation and applies it to new observations with different fundamental frequencies. This is achieved by resampling each observation to match the time scale of the reference signal prior to inference.

Simulation results demonstrate that the performance of the R-DCNN is comparable to that of classical methods, including DCNN and AR approaches, which involve significantly higher computational complexity due to per-observation retraining. This makes the proposed method a promising alternative for periodic signal denoising in power-limited environments.

A key limitation of the method is the requirement for accurate knowledge of the fundamental frequency for each observation. Future work will focus on evaluating the robustness of the method to errors in fundamental frequency estimation and resampling precision. This analysis is essential, as achieving high accuracy in these parameters may introduce additional computational overhead, which must remain minimal in practical applications.

The authors welcome collaboration to evaluate the proposed method across different applications.

**Declaration of competing interest**

The authors declare that they have no known competing financial interests or personal relationships that could have appeared to influence the work reported in this paper.